\documentclass[letterpaper, 10 pt, conference]{ieeeconf}  

\IEEEoverridecommandlockouts                              

\usepackage{graphicx}
\usepackage{adjustbox}
\usepackage{dblfloatfix}
\usepackage{amsmath}

\title{\LARGE\bf
HRI-SA: A Multimodal Dataset for Online Assessment of Human Situational Awareness during Remote Human-Robot Teaming
}

\author{Hashini Senaratne$^{1}$, Richard Attfield$^{2}$, Samith Widhanapathirana$^{1}$, David Howard$^{1}$, Cecile Paris$^{1}$, \\Dana Kulic$^{1,2}$, Leimin Tian$^{1}$
\thanks{$^{1}$CSIRO Robotics: www.csiro.au/en/research/technology-space/robotics}
\thanks{$^{2}$Monash Robotics: www.monash.edu/engineering/robotics}
}

\begin{document}

\maketitle
\thispagestyle{empty}
\pagestyle{empty}

\begin{abstract}

Maintaining situational awareness is critical in human–robot teams. Yet, under high workload and dynamic conditions, operators often experience situational awareness gaps. 
Automated detection of situational awareness gaps could provide timely assistance for operators. However, conventional situational awareness measures either disrupt task flow or cannot capture real-time fluctuations, limiting their operational utility. To the best of our knowledge, no publicly available dataset currently supports the systematic evaluation of online human situational awareness assessment in human-robot teaming. To advance the development of online situational awareness assessment tools, we introduce HRI-SA, a multimodal dataset from 30 participants in a realistic search-and-rescue human-robot teaming context, incorporating eye movements, pupil diameter, biosignals, user interactions, and robot data. The experimental protocol included predefined events requiring timely operator assistance, with ground truth situational awareness latency of two types (perceptual and comprehension) systematically obtained by measuring the time between assistance need onset and resolution. We illustrate the utility of this dataset by evaluating standard machine learning models for detecting perceptual situational awareness latencies using generic eye-tracking features and contextual features. Results show that eye-tracking features alone effectively classified perceptual situational awareness latency (recall=88.91\%, F1=67.63\%) using leave-one-group-out cross-validation, with performance improved through contextual data fusion (recall=91.51\%, F1=80.38\%). This paper contributes the first public dataset supporting the systematic evaluation of situational awareness throughout a human-robot teaming mission, while also demonstrating the potential of generic eye-tracking features for continuous perceptual situational awareness latency detection in remote human-robot teaming.
\end{abstract}

\section{INTRODUCTION}

Human operators' Situational Awareness (SA) is a critical determinant of performance and mission success in human–robot teams~\cite{shayesteh2022enhanced,wong2017workload, doi:10.1177/154193120805200410}. As robotic systems continue to advance in capability and the scale of human–robot teams expands, maintaining required levels of SA for operators has become increasingly challenging~\cite{10.1145/3588325, roldan2017multi}. This challenge is particularly evident in field robotics, where one human often remotely guides a fleet of semi-autonomous robots under high workload, environmental uncertainty and dynamic conditions. Examples include the DARPA Subterranean Challenge~\cite{10.1145/3588325}, multi-Unmanned Aerial Vehicle (UAV) missions~\cite{cummings2007automation}, and agricultural robotics~\cite{senaratne2023measuring}. A recent interview study identifies recurring and frequent SA gaps in such contexts, reflecting intermittent misalignment between operators' SA and the task- and time-specific SA requirements~\cite{10.1145/3743693}. These SA gaps lead to poor mission performance, heightened risks to robots, bystanders and the environment, and increased operator frustration and stress. According to this study, operators expressed a strong preference for robotic systems that actively support the mitigation of SA gaps. 

Currently, most robotic systems offer limited or no support for operators to maintain or regain required levels of SA~\cite{10.1145/3743693}. Measuring SA during Human-Robot Interaction (HRI) is crucial for meaningful assistance.  Traditionally, SA has been assessed using freeze-probe methods (e.g., SAGAT: the Situation Awareness Global Assessment Technique) and post-trial subjective measures (e.g., SART: Situational Awareness Rating Technique)~\cite{hopko2021effect,dini2017measurement,endsley1988situation}. While these approaches are well-established in controlled research environments, they require interruptions to task flow, do not capture fine-grained SA fluctuations and are subject to recall, overgeneralization and reactivity biases~\cite{endsley2000situation}. Therefore, these measures are unsuitable for real-time HRI scenarios~\cite{ali2025estimating, 10.1145/3743693}. 

The advancement of automated and online SA gap assessment in HRI is fundamentally constrained by the absence of publicly shared, systematically annotated datasets. Prior studies have explored quantitative proxies of SA, predominantly through probe-based techniques and indirect performance metrics, and only occasionally incorporating informative eye-tracking or physiological signals~\cite{steinfeld2006common, cai2025assessing,dini2017measurement,ali2025estimating,ratwani2010single,senaratne2023measuring}. However, these datasets are rarely released and typically remain inaccessible beyond the original research teams. Consequently, the broader research community lacks benchmark data to systematically evaluate, compare, and replicate methods for detecting SA gaps in an online manner. To the best of our knowledge, to date, no openly available dataset enables rigorous evaluation of continuous SA gap detection across an entire HRT mission. This limited transparency and reproducibility restrict methodological progress, impede fair comparisons across approaches, and slow both the discovery of new SA assessment techniques and the translation of promising methods into operational HRT systems.

One frequent and operationally consequential SA gap in HRT is SA Latency (SAL)—a delay in recognizing a robotic system's need for assistance or changes in the mission-relevant environment~\cite{10.1145/3743693}. The ability to detect SAL in real-time could meaningfully aid operators to meet dynamic SA demands during complex HRT missions. A previous study reported strong correlations between certain eye-tracking features and two types of SALs: perceptual and comprehending, but it involved only a single participant and analyzed only SAL durations rather than the continuous mission timeline~\cite{senaratne2023measuring}. To our knowledge, no dataset from realistic HRT scenarios currently exists and supports continuous, mission-wide assessment of SAL, and no prior work has conducted such an evaluation.

To enable the development of online approaches for SA assessment, this paper contributes a multimodal dataset collected continuously during a realistic search-and-rescue HRT experiment, where various SAL situations occur. Specifically, events requiring operator assistance were systematically embedded based on the map location, yet the occurrence and timing of SA latency emerged spontaneously and differed across participants, depending on the robots' arrival at each location, their direction and order of approach and the operator’s responsiveness. We illustrate the utility of this dataset by evaluating a set of Machine Learning (ML) models for classifying the presence and absence of Perceptual SAL (PSAL) every 5 seconds throughout the experiment. The key contributions of this paper are:

\begin{itemize}
\item HRI-SA\footnote{The access link will be included in the published version of this paper.}, a novel multimodal dataset of a human-robot teaming task, collected from 30 participants with varying expertise, for evaluating methods for assessing human SA gaps.  The dataset consists of operator eye-tracking data, biosignals, interaction data, and robot data, along with ground truth labels for perceptual and comprehending SA latencies during specific moments that required different types of operator assistance (waypoints, teleoperation, verifying robot detected objects).
\item Demonstration that generic eye-tracking data effectively classifies Perceptual SA Latency (PSAL) in an online manner, with performance enhanced by the fusion of minimal and readily available contextual information. 
\end{itemize}


\section{Related Work}

Situational awareness has been defined using three levels: perception (recognizing elements in the environment), comprehension (understanding their meaning), and projection (anticipating their near future status)~\cite{endsley1988situation}. Examples of perceptual and comprehending SA demands in HRI include timely recognition of robot errors or interface notifications, and detecting non-optimal robot behavior, respectively~\cite{senaratne2023measuring}.

The ability to measure operator SA in real-time would enable the system to alert the operator to provide timely interventions only when needed, reducing alert fatigue and improving team performance~\cite{10.1145/3743693,biswas2024modeling}. It would also allow robots to adapt and coordinate effectively with humans. However, subjective self-assessment and probing techniques commonly used in Human-Computer Interaction (HCI) and HRI research are unsuitable for real-time applications~\cite{10.1145/3743693,schuster2011measurement}. These measures often require freezing the task, which may not be possible in real-world HRI scenarios, and they may also be prone to bias, especially if administered after a long mission. In addition, they fail to capture SA at a high temporal resolution; i.e., they do not specify when SA changes occur and how they relate to specific tasks. They provide only an overall SA score, typically obtained after a mission or at limited time points. These limitations highlight the need for objective, continuous measures of operator SA. However, research on objective online SA assessment in HRT remains sparse, and comprehensive datasets enabling the systematic discovery and validation of such measures are largely nonexistent, as reviewed below.

\subsection{Contextual Data Based Human SA Assessments}

Task-level or event-based performance measures derived from contextual data are often used to assess operator SA. For example, increased task completion times and a higher number of mistakes per task (e.g., incidents of UAVs colliding with hazards and contact errors during manipulation tasks) can be proxy indicators of poor SA~\cite{ratwani2010single, steinfeld2006common}. Although these measures may correlate with SA, they can also be influenced by other factors; e.g., robot sensing and data integration~\cite{10.1145/3743693}. Furthermore, calculating these measures can be infeasible for complex and dynamic HRT missions with unpredictable task durations, overlapping tasks, and flexible workflows. Additionally, many of these measures depend on specific robotic system data and context, making generalization across applications challenging.

\subsection{Eye-tracking Data Based Human SA Assessments}

Eye-tracking data provide a non-intrusive, detailed view of operator attention allocation, information processing and cognitive load~\cite{rayner1998eye,beatty1982task}. Eye-tracking can be done continuously in real-time without disrupting task flow. Additionally, metrics derived from eye-tracking data can be integrated with general contextual data about robots, their environment and interface (e.g., areas of interest in an interface, robot pose), enabling SA estimation across various HRT applications.

Eye-tracking metrics that are informative for assessing operator SA include fixation points (i.e., stable locations where the eyes are focused), fixation duration, saccade patterns (rapid eye movements between fixations), and pupil diameter~\cite{zhang2023physiological}. Features derived from these metrics—e.g., mean fixation duration on an Area of Interest (AOI) or latency to first fixation on a critical element—provide quantitative indicators of attention allocation and information uptake~\cite{senaratne2023measuring}. When mapped to mission-relevant events or system states, such features can be translated into estimates of SA, particularly at the perceptual and comprehension levels.

Many HCI studies~\cite{de2019situation,van2012eye} and HRI studies with a single, limited mobility robot~\cite{cai2025assessing} rely on fixed AOIs for deriving eye-tracking features. While fixed AOIs—e.g., around notification panels—can be useful for detecting certain perceptual SA gaps, they transfer poorly to multi-robot and dynamic HRI settings. Only a few studies have examined dynamic AOIs for moving entities (e.g., aircrafts,  robots) to evaluate operator SA using eye-tracking features~\cite{moore2010development,senaratne2023measuring}.

Several HCI studies~\cite{zhang2023physiological} and a few HRI studies (e.g.,~\cite{cai2025assessing,dini2017measurement}) report differences in eye-tracking features between high- and low-SA levels. However, most do not capture fine-grained and precise SA states, as their ground truths are derived from overall task performance, post-study questionnaires, or infrequent probing~\cite{cai2025assessing, jiang2021correlation, dini2017measurement,de2019situation,van2012eye}. 



Few studies have examined SA at specific intervention events in HRT missions, where comprehending SALs are likely to emerge. For example, in a reconnaissance mission simulation with two Unmanned Ground Vehicles (UGVs), Ali et al.~\cite{ali2025estimating} classified SA as high or low based on the occurrence of help requests at intervention points, achieving 74\% accuracy. Similarly, Ratwani et al.~\cite{ratwani2010single} studied the supervision of five UAVs, categorizing SA based on whether or not the operator acted in time to prevent hazard collisions, reporting an F1 score of 62.79\% (recall=84.37\%, precision=50\%). Both used logistic regression models with eye tracking and secondary task interaction metrics. Ratwani et al.’s model is context-specific and less applicable to unknown environments because it required pre-marked hazards and targets on a map. Notably, both studies analyzed data just preceding a SAL event but not during the latencies, which is essential for realistically classifying the presence of SALs. Also, both used 10-fold cross validation, which may overestimate performance due to overlap in datapoints from the same participant between training and test sets~\cite{lumumba2024comparative}.

In contrast, Senaratne et al.~\cite{senaratne2023measuring} evaluated the correlation between perceptual and comprehending SALs and eye-tracking features during a two-robot smart farm mission. PSAL was measured by the operator’s delay in verifying robot-detected fruits, while Comprehending SAL (CSAL) was measured from the delay in identifying non-optimal robot behavior. Using robot-specific dynamic AOIs and static AOIs for interface panels, they found PSAL correlated strongly with mean off-task fixation time and off-task saccade duration, whereas CSAL correlated with the 
percentage of saccade duration in relevant AOIs. However, these findings are limited by its single-participant sample and consideration of only time windows around SAL events.

More broadly, existing limited research related to SA gaps has focused on higher-level comprehending SA, with less attention to perceptual SA. Yet perceptual SA underpins the SA process, making it a principled starting point for online SA gap assessment. Moreover, existing studies lack analysis on instances where robots perform optimally and no operator intervention is needed~\cite{senaratne2023measuring,ratwani2010single,ali2025estimating}. This exclusion of optimal cases may bias performance estimates and affect the suitability of findings for real-time applications. 

\begin{figure}[t]
    \centering
    \includegraphics[width=0.95\linewidth]{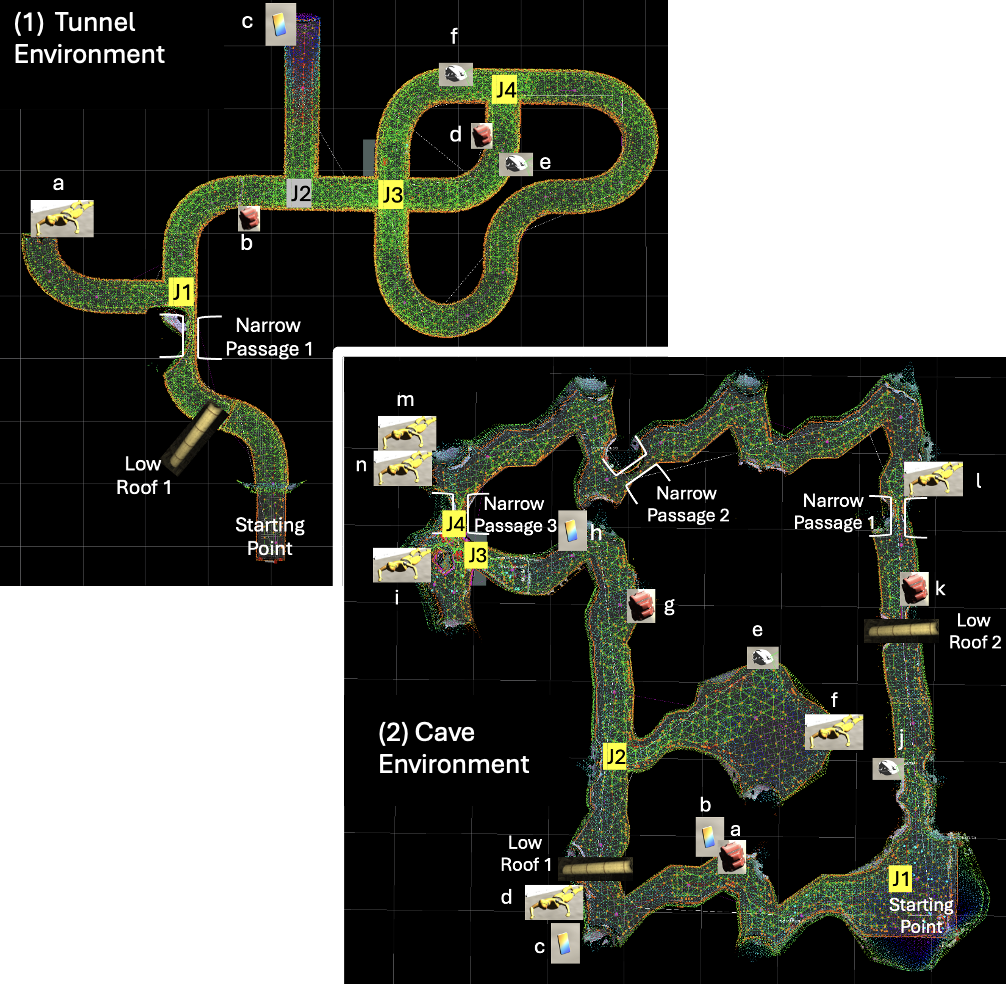}
    \caption{\centering{Two environments used in the user study: Tunnel and Cave. The robot required teleoperation assistance at low roofs and narrow passages, and waypoint guidance at intersections marked in yellow. If the robot explored the tunnel efficiently, it would detect objects in the order of a-f. In the cave, one robot would detect objects a-i while the other objects j-n.}}
    \label{fig:two_maps}
\end{figure}

\subsection{Data Availability}
To date, none of the above reviewed studies provide publicly available datasets with systematic annotations of SA or SA gaps in HRI (e.g., SAL). Existing HRI datasets have been collected for other purposes and lack ground truth labels for continuous or event-specific SA states. Consequently,  researchers cannot benchmark, replicate, or systematically evaluate methods for online SA assessment in HRT.

\section{HRI-SA Dataset}

To address the gaps identified above, we conducted a user study in a simulated search-and-rescue HRT mission to collect continuous SAL and multimodal data, i.e., the HRI-SA dataset, as detailed in this section. 

\subsection{Human-Robot Teaming Experiment Setup}

The HRT mission was to explore two disaster sites and quickly detect signs of casualties. These two experimental conditions were counterbalanced across subjects. The tunnel, a simpler environment explored with one robot, was expected to yield shorter SALs, while the complex cave, explored with two robots, aimed to cause longer SALs (see Figure~\ref{fig:two_maps}). Both environments featured low roofs and narrow passages, making navigation difficult for robots without human teleoperation assistance at these points. To optimize map exploration, the operator also had to monitor and occasionally override robots' decisions at intersections by issuing way-points.

Robots were equipped with RGB cameras and a spinning lidar and could autonomously explore and map the unknown disaster site using a multi-agent SLAM system and an automated task allocation system~\cite{kottege2024heterogeneous}. As in the DARPA Subterranean Challenge~\cite{10.1145/3588325}, the robots also detected signs of casualties and communicated progress to other robots and via the supervisory interface.

\begin{figure*}[t]
    \centering
    \includegraphics[width=0.99\linewidth]{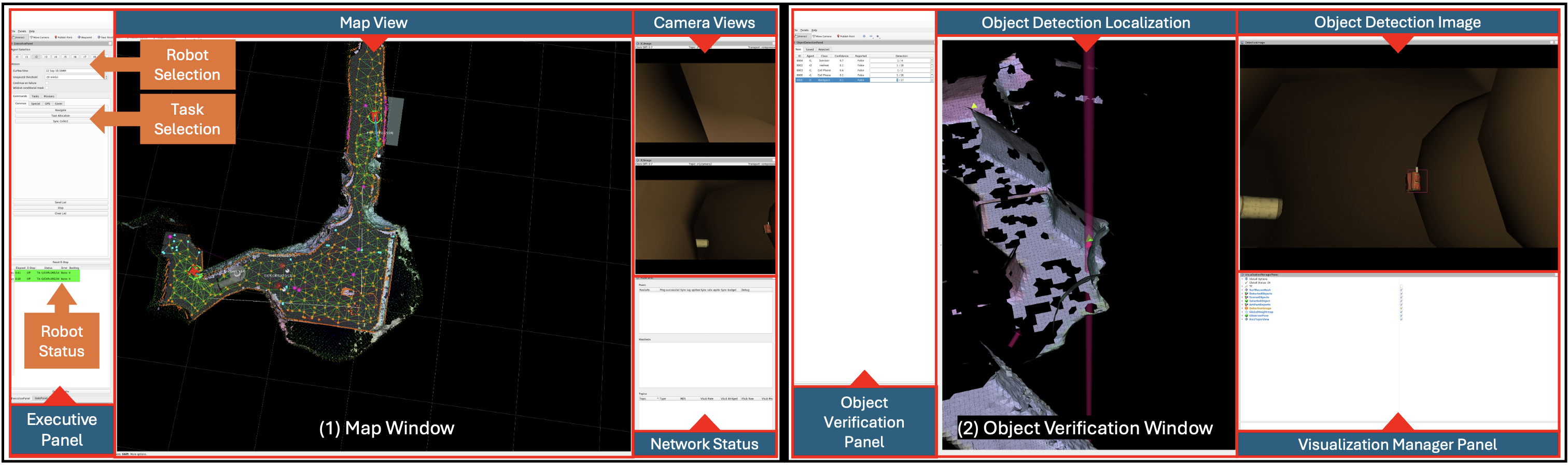}
    \caption{\centering{Graphical user interface used during the user study. The operator can (1) view robots’ status, actions and their operating environment and (2) view and verify the robot-detected signs of casualties and their localization. One window is viewable at a time.}}
    \label{fig:user_interface}
\end{figure*}

Apart from monitoring the robots and performing supervisory control as needed, the operator's role was to verify robot-detected signs of casualties (survivors, backpacks, helmets and cellphones). The operator interface included: (1) a live 3D map with robot models, their task markers, and planned trajectories; (2) front and rear camera feeds from the robots; (3) an executive panel for selecting, stopping or tasking robots; (4) robot-status panels; and (5) a separate window to inspect detected objects (see Figure~\ref{fig:user_interface}). The operator could issue waypoint commands on the 3D map and use a joystick controller to teleoperate a selected robot.

\subsection{Participants}

The study was reviewed and approved by an institutional ethics committee. Thirty participants (12 women, 18 men; age = [M = 32.52 (9.12)]) were recruited via invitations shared through a research institution and a university, and through snowballing. All had previously used a joystick controller for video games or robotic systems, with self-identified expertise varying from novice (5), advanced beginner (4), competent (10), proficient (7), to expert (4). Among them, 22 had prior experience with a robotic Graphical User Interface (GUI), with varied expertise: novice (5), advanced beginner (3), competent (8), proficient (3), and expert (3).

\subsection{Study Procedure and Subjective Data}

Upon volunteering, participants provided informed consent and completed an online questionnaire to report demographic details and prior HRT experience.

Each study session lasted approximately 90 minutes. Participants first donned a smartwatch, which measured their biosignals (details in Section~\ref{sec:sensor}) and displayed time. They then completed a tutorial with instructional videos explaining the team's mission, intervention timing, user interface, and interaction mechanics. Between videos, they practiced interactions in a demo environment, including teleoperating a robot through a low roof and a narrow passage, stopping a robot, providing waypoints, enabling automated exploration mode on a robot, navigating the map view, viewing multiple images linked to object detection and verifying detections. The tutorial also provided guidance on how to supervise robots at intersections. Participants were instructed to teleoperate only if the robot struggled. They were further instructed to wait until robots captured five images of the same object before rejecting an entry if no sign of casualty appeared in the first set of images sent for verification. 

After the tutorial and practice, a questionnaire was administered to assess participants' understanding of the instructions. The experimenter clarified the relevant instructions for any incorrect response. Participants received a printed cheat-sheet for reference during the experiment if needed. Before the experiment, the eye tracker was calibrated for the participant (see Section~\ref{sec:sensor}).

During the experiment, participants collaborated with robots in two environments. In the cave, two robots explored in parallel, but only one robot's camera feeds were shown, aiming to increase SA latencies in guiding the other robot. A stratified randomization technique assigned participants to environment orders based on their expertise (expert vs. novice, defined by prior system experience). Those who had not reached all objects in the first environment by the end of the first hour, but had spent at least 10 minutes on that run, were instructed to stop and proceed to the next mission. Two participants (P4 and P19) spent over an hour on pre-experiment activities, leaving time to complete only one run.

Each run included pre-placed events and configurations designed to elicit SALs (see Figure~\ref{fig:two_maps}), occurring at different times across runs depending on robot arrival, approach order and direction, and operator responsiveness. Participants switched between two interface windows to identify and respond to events, by reviewing robot-supplied object images, teleoperating robots struggling at low roofs or narrow passages, and issuing waypoints when robots took incorrect turns at intersections. They were asked to verify objects as soon as possible, at most within 1 minute of detection.

After each run, participants completed a questionnaire ranking events related to object verification, teleoperation, waypoint issuance and monitoring, focusing on self-perceived timing for intervention recognition and mental demand. Open-ended questions gathered instances of delayed or excessive awareness, confusion, stress, boredom, and varying cognitive load. Following both runs, participants reported preferred assistance methods to minimize delays in recognizing the need to act in remote HRT. These subjective data are de-identified and included in the dataset, with selected quotes presented in Section \ref{sec:discussion} to suggest potential ways this dataset can be leveraged for assessing online SA gaps and applied meaningfully in future.

Sample videos from both runs, snapshots of the tutorial and the administered questionnaires are provided in the supplementary materials. See: https://youtu.be/fV5XzLTKOsk.

\subsection{Sensors and Objective Data}\label{sec:sensor}

To track participants' eye movements and pupil diameter during the two runs, we used a Tobii Pro Fusion eye tracker mounted below a 31.5-inch screen and configured for 120 Hz sampling rate (0.3° accuracy). Participants sat about 60 cm away from the display. A five-point calibration was performed with the Tobii Pro Eye Tracker Manager after the practice session and repeated before the final run if the participant's seating position had shifted. We developed a Robot Operating System (ROS) node to save incoming and processed data from the eye tracker to two data files: (1) raw x and y gaze positions, pupil diameter, and validity metrics for each eye, and (2) whether the participant’s gaze is within 50 px of the positions of robot 1 or robot 2 on the map panel.

We also saved robot and interface events—including robot poses (1 Hz), waypoint commands, joystick inputs, other control commands (stop, automatic exploration), object detections, and object verification actions (select, accept, reject)—as timestamped ROS bag files. A separate Python script logged keyboard events to track switches between the map and object verification windows via Alt and Tab keys.

Further, we collected participants' electrodermal activity, photoplethysmogram, skin temperature, and wrist movement data with an EmbracePlus smartwatch worn on their non-dominant hand. Speech was recorded using a smartphone on the table. While the analysis in this paper uses only eye-tracking data, our publicly released dataset includes all the collected biometrics, excluding speech for privacy reasons.


\subsection{Synchronization}\label{sec:sync}

Objective data was collected on a computer synchronized to the Network Time Protocol (NTP) time, with each entry timestamped. An additional synchronization step was required for the robotic system and keyboard event data due to observed clock drift in the robotic system. It aligned object verification accept and reject events with corresponding keyboard events, calculated the drift rate using the first and last verification, and adjusted the timestamps accordingly.

\subsection{Ground Truth Labels}\label{sec:groundtruth}

PSAL was evaluated on the task of verifying signs of casualties. This task closely relates to perceptual SA, as recognizing a new object for verification was a perceptual task, since it appeared as an item on a list. The PSAL for an object entry was determined as \( t_{iv} - t_{sv} \), where \( t_{iv} \) is the time at which the operator initiated the verification, and \( t_{sv} \) is the time at which the object was submitted for verification. If a verification was not initiated, the run’s end time was used for \( tiv \). The PSAL experienced by the operator was determined for every 5 seconds throughout each run, as follows: \[
PSAL_{now} = \max_{1 \leq i \leq n} (t_{now} - t_{i})
\]
Here, \( n \) is the number of objects in the list for the operator to initiate a verification, \( t_{now} \) is the current time, \( t_{i} \) is the time at which the $i^\text{th}$ object was submitted for verification. 



CSAL was quantified based on the operator's ability to identify situations where robot performance degraded, including requiring teleoperation to navigate low roofs or narrow passages and waypoint assistance for efficient exploration. Recognizing these situations required the operator to synthesize information from robot poses, map views (via rotation and zooming), and multiple interface panels (e.g., map and camera views). The CSAL was calculated as \( t_{ar} - t_{no} \), where \( t_{ar} \) is the time when the operator indicated awareness of the event (by initiating teleoperation or stopping the robot) or the time when the event was resolved without being noticed by the operator, and \( t_{no} \) is the time when the robot began to act non-optimally. For this, robotic system data (robot pose, map coordinates, joystick inputs, waypoint and stop commands) were used. The ground truths provided with the dataset are based on the following criterion, though alternative thresholds can be applied: 
if a robot moved 10 distance units past an intersection, narrow passage, or low roof without operator assistance, the need for intervention was considered resolved, as the robots' automated task allocation system will begin other tasks. 
The $CSAL_{now}$ variable was calculated every 5 seconds for each run and robot, similar to the $PSAL_{now}$, where n is the number of different interventions required by a robot and \( t_{i} \) is the time at which the $i^\text{th}$ intervention was first required. 


\subsection{Data Preprocessing}

\subsubsection{Data Formatting and Cleaning} After converting ROS bag files, eye-tracking and key logger data files to .csv format, corrections were made for missing column names and split columns caused by extra delimiters. Next, the synchronization steps were performed (see Section~\ref{sec:sync}). 

\subsubsection{Data Discards} Key logger data was entirely lost for P4, P7 and P19 in the tunnel environment and partially lost for P1 at the end of two runs. Eye-tracking data was missing for P23 and mostly invalid for P13 in the cave environment. 

\subsubsection{Pre-processing Gaze Events}
Key logger events were utilized to identify the screen being viewed (object verification or map). This data was combined with the robot of focus to determine four AOIs for eye movement events: object verification, robot 1-related, robot 2-related, and away. The robot-related AOIs are dynamic; e.g., if an operator viewed robot 1’s camera feed and then focused on robot 1 or other areas of the map before attending to robot 2, all eye movement events in this period pertain to the ``robot 1-related'' AOI. Eye movements outside the screen boundaries were labeled as looking ``away.''

To classify eye movement events into fixations and saccades, we implemented the Velocity-Threshold Identification (aka I-VT) filter in Python, following~\cite{olsen2012tobii}. This filter offers several configuration options; we selected those suitable for the HRI-SA dataset. First, noise was reduced using a moving median filter with a 3-sample window. Next, data from both eyes were merged by averaging valid x and y coordinates for each event. Unlike the Tobii I-VT filter, which uses average velocity per sample, we computed gaze velocity (degrees / second) between consecutive eye movement events, as follows, where the function deg() converts radians to degrees. This method increased the sensitivity to rapid onsets of saccades or fixations.


{\fontsize{7}{9}\selectfont
\[
v = 
\frac{
\sqrt{
\left( \deg \bigl( \arctan \tfrac{\text{gaze\_x\_cm}}{\text{viewing\_distance\_cm}} \bigr) \right)^{2}
+
\left( \deg \bigl( \arctan \tfrac{\text{gaze\_y\_cm}}{\text{viewing\_distance\_cm}} \bigr) \right)^{2}
}
}{
{\text{time\_difference}} 
}
\]
}

We used the Tobii I-VT filter's default velocity threshold of 30°/s to classify events as fixations (below) and saccades (at or above). Adjacent eye movement events were merged based on a maximum angle of 0.5° and a maximum duration of 75 milliseconds between gaze events, following prior research~\cite{komogortsev2010standardization}, without merging across distinct AOIs. Fixations shorter than 60 ms (default minimum fixation duration used in Tobii) were labeled unclassified due to their very brief, potentially erroneous, or physiologically impossible nature. 

The HRI-SA dataset includes the raw and preprocessed data described above, PSAL and CSAL ground-truth labels, and the eye-tracking features described below.

\section{Example Dataset Utility: PSAL Detection}

Here, we illustrate the utility of our HRI-SA dataset by evaluating standard ML models for detecting PSALs using generic eye-tracking features and contextual features. After discarding the affected runs or time periods, 28 participants remained for this analysis.

\subsection{Feature Extraction}

Six eye-tracking features were extracted every 5 seconds for each run. Five (F1-F5) were derived for the intervals of 10 s, 20 s, ..., up to 180 s prior to the considered timestamp, and another (F6) was derived for the considered timestamp, yielding 91 features (18 × 5 + 1) per timestamp. Additionally, two contextual features were extracted from immediate robotic system data for each timepoint. 

F1: Percentage of time fixating on relevant AOIs. Here, \textit{j} and \textit{n} are the first and last eye movement event indices within the considered period with duration \textit{T}, and \(t_{{fixationAOI_{i}}}\) is the fixation duration for the $i^{\text{th}}$ event hitting the relevant AOI (i.e., object verification AOI for PSAL analysis).
   \[
   \frac{\sum_{i=j}^n t_{{fixationAOI_{i}}}}{T} * 100
   \]

F2: Mean off-task fixation time. Here, \textit{j} and \textit{n} are as for F1, \(t_{{fixationOFFAOI_{i}} }\) is the fixation duration for the $i^{\text{th}}$ event hitting an irrelevant AOI (robot 1-related, robot 2-related, or away), and \textit{N} is the number of fixations on an irrelevant AOI.
   \[
   \frac{\sum_{i=j}^{n} t_{{fixationOFFAOI_{i}}}}{N}
   \]

F3: Percentage of saccade duration on relevant AOIs, calculated similarly to F1 but using saccade durations.

F4: Mean off-task saccade duration, computed similarly to F2 but using saccade durations.

F5: Mean off-task pupil diameter, calculated similarly to F2 but using pupil diameter data for eye-tracking events.

F6: Whether currently viewing a relevant AOI.


F7: Number of objects pending verification.

F8: Whether any objects remain for verification (0 or 1).



These features are chosen as generic and derivable from immediate data, making them applicable across diverse HRT systems. For example, object verification events are mappable to error notifications or robot help requests. The above extracted features are also included in the dataset.

\subsection{Analysis Method for Classifying PSAL}

Ten ML models were implemented to classify the presence or absence of PSALs using three feature sets: (1) eye-tracking, (2) contextual, and (3) a combined set of both. The models included: classifiers assuming linear decision boundaries: Logistic Regression (LR) and Linear Discriminant Analysis (LDA); nonlinear probabilistic models: Quadratic Discriminant Analysis (QDA) and Naive Bayes (NB); an instance-based model: K-Nearest Neighbors (KNN), which compares new samples to training data using a similarity measure; tree-based models: Decision Tree (DT), Random Forest (RF), and AdaBoost, which partition the feature space using hierarchical rules; a kernel-based method: Support Vector Machine (SVM), enabling flexible decision boundaries; and a neural network: Multi-Layer Perceptron (MLP), with interconnected layers of artificial neurons for hierarchical feature learning. These standard ML models were adopted to evaluate model suitability for different feature sets and feature importance for PSAL detection. All models were developed in Python using the Scikit-learn library, and hyperparameters were optimized using GridSearchCV. 

The presence or absence of PSAL at time $t_{now}$ was determined as follows, with variables defined in Section \ref{sec:groundtruth}. We chose a 30 seconds threshold: i.e., latency values exceeding 30 seconds were labeled as PSAL, and shorter ones as no PSAL. 
Since participants were instructed to verify detected casualties as soon as possible and no later than one minute, detecting PSALs beyond 30 seconds was considered useful for issuing early alerts. 
There were 1,155 PSAL datapoints, and 6,071 for no PSALs.


 \[
  IS\_SAL_{now} = \mathbf{1}\!\left( \max_{1 \leq i \leq n}  (t_{now} - t_{i}) >= threshold \right)
 \]

To tackle class imbalance in the dataset, we used the \textit{class\_weight} \textit{=`balanced'} setting where applicable, enabling algorithms adjust class weights inversely proportionate to class frequencies, thus enhancing sensitivity to the minority class. For models without this setting, we adjusted class weights at the base estimator level, specified class priors when feasible, or used SMOTE (Synthetic Minority Over-sampling Technique) to up-sample the minority class~\cite{chawla2002smote}.  

\begin{table*}[b!]
\centering
\caption{Perceptual Situational awareness latency classification performance of machine learning algorithms across feature groups}
\resizebox{\textwidth}{!}{%
\begin{tabular}{|l|ccc|ccc|ccc|}
\hline
{ML} & 
\multicolumn{3}{c|}{Contextual Features} & 
\multicolumn{3}{c|}{Eye Features} & 
\multicolumn{3}{c|}{Eye + Contextual Features} \\ \cline{2-10} 
Model & F1 Score \% & Recall \% & Precision \% 
 & F1 Score \% & Recall \% & Precision \% 
 & F1 Score \% & Recall \% & Precision \% \\ \hline
LR 
& 57.34 (14.65) & \textbf{100.00 (0.00)} &  41.68 (14.56)
& 44.62 (15.48) & 81.69 (19.51)  & 31.94 (13.24) 
& 71.68 (11.25) & 94.45 (5.79) &  58.90 (13.13) \\
LDA 
& 57.34 (14.65) & \textbf{100.00 (0.00)} &  41.68 (14.56)
& 40.92 (11.80) & \textbf{90.77 (8.09)} & 27.17 (10.04) 
& 64.09 (11.83) & \textbf{98.47 (2.16)} & 48.58 (12.76) \\
QDA 
& 57.34 (14.65) & \textbf{100.00 (0.00)} &  41.68 (14.56)
& 44.90 (13.23) &  88.24 (10.37) & 31.08 (11.55) 
& 71.32 (11.64) & 93.38 (6.99) & 58.56 (12.96) \\
NB 
& 57.34 (14.65) & \textbf{100.00 (0.00)} &  41.68 (14.56)
& 41.59 (12.01) & 90.67 (8.03) & 27.74 (10.24) 
& 72.73 (11.06) & 91.74 (7.34) & 61.19 (12.92) \\
KNN 
& 62.12 (17.05) & 70.50 (21.23) &  58.89 (18.88)
& 62.14 (16.03) & 70.00 (21.18) & \textbf{59.27 (17.37)}
& 67.63 (11.68) & 88.91 (8.94) & 56.76 (15.43) \\
DT 
& 57.34 (14.65) & \textbf{100.00 (0.00)} &  41.68 (14.56)
& 54.75 (15.55) & 86.22 (8.07) & 43.31 (19.08) 
& 70.26 (14.33) & 86.41 (16.37) & 61.50 (14.93) \\
RF 
& 57.34 (14.65) & \textbf{100.00 (0.00)} &  41.68 (14.56)
& 55.66 (14.65) & 81.67 (10.85) & 45.58 (18.38)
& \textbf{80.38 (10.87)} & 91.51 (5.97) & \textbf{72.89 (14.48)} \\
AdaBoost 
& 57.34 (14.65) & \textbf{100.00 (0.00)} &  41.68 (14.56)
& 54.89 (14.33) & 86.58 (6.71) & 41.81 (14.66) 
& 79.29 (10.47) & 92.01 (7.80) & 71.14 (13.82) \\
SVC 
& 57.34 (14.65) & \textbf{100.00 (0.00)} &  41.68 (14.56)
& 47.55 (13.51) & 79.76 (12.60) & 35.36 (13.03) 
& 74.70 (10.70) & 93.93 (4.94) & 63.01 (13.03) \\
MLP 
& \textbf{71.65 (12.25)} & 74.19 (21.51) &  \textbf{67.13 (18.19)}
& \textbf{67.63 (11.68)} & 88.91 (8.94) & 56.76 (15.43)
& 73.22 (12.45) & 94.07 (5.34) & 61.57 (15.38) \\
\hline
\end{tabular}%
}
\label{tab:results}

\end{table*}


We employed cross-validation for hyperparameter tuning and model evaluation. As the commonly used leave-one-participant-out strategy resulted in some test folds without positive instances, preventing meaningful performance evaluation, we adopted a leave-one-group-out cross-validation scheme~\cite{yates2023cross}, grouping two participants based on their contributions to PSAL datapoints (i.e., ranking the participants by the number of positive samples and pairing the highest with the lowest, repeating until all were paired). This resulted in 14 groups (28/2), ensuring adequate class representation in each test fold and allowing evaluation of two new individuals.

Model performance was evaluated using Precision, Recall, F1-score, mean Area Under the Curve (AUC). Accuracy was excluded, as class imbalance can inflate it through majority-class predictions, despite poor minority-class predictions.





\subsection{Results: PSALs Classification Performance}

To minimize false negatives (PSALs misclassified as no latency) and false positives (no latency events misclassified as PSAL), we selected the best model based on F1 scores. 

Considering unimodal features, the MLP model performed best for classifying PSALs (see Table~\ref{tab:results}), achieving an F1 score of 67.63\% (recall=88.91\%, precision=56.76\%) with only eye-tracking features and 71.65\% (recall=74.19\%, precision=67.13\%) with only contextual features. For the combined feature set, the RF model outperformed others with an F1 score of 80.38\% (recall=91.51\%, precision=72.89\%). Mean AUC across all folds for these were 0.77 (SD=0.08), 0.90 (SD=0.06), and 0.97 (SD=0.02), respectively. 

We analyzed the RF model with the highest F1 score to identify relative feature importance for PSAL  detection. The key features were two contextual—F8 (0.23) and F7 (0.20)—and three eye-tracking features based on saccades, fixations and pupil diameter extracted from 30-40 seconds prior—F3\_40 (0.06), F2\_40 (0.05) and F5\_30 (0.05).

\section{Discussion}\label{sec:discussion}

We presented HRI-SA, a multimodal dataset for investigating Situational Awareness (SA) gaps during remote HRT. Along with eye-tracking and physiological data, it provides ground truth annotations of perceptual and comprehending SA Latencies (SALs). To the best of our knowledge, this is the first dataset offering realistic mission-wide data on SA, covering time points where human intervention is needed as well as not. It also includes subjective reports of excessive awareness, confusion (i.e., incomplete awareness), stress, boredom, and cognitive load, enabling investigation of SA gaps and user states beyond SAL~\cite{10.1145/3743693}.

However, this dataset has limitations. It was collected during simulated missions, where operators may exhibit lower alertness than in physical missions, due to lower risks. Also, simulated environments are less complex than real-world settings. Thus, the eye-tracking data may not fully represent real-world HRT. The dataset also provides ground truths or data useful for deriving ground truths for Perceptual SAL (PSAL) and Comprehending SAL (CSAL) only for specific tasks (but transferrable to other contexts) and does not account for predictive SALs (i.e., delays in predicting the robot's near-future needs), which are challenging to capture without delving into the operators' thought processes.

In addition to the primary dataset contribution, we illustrate its utility through a preliminary ML experiment investigating the efficacy of eye-tracking and contextual features for PSAL classification. Generic eye-tracking features achieved moderate performance (F1=67.63\%) in classifying whether or not an operator is experiencing a PSAL, comparable to contextual features alone (F1=71.65\%). Combining feature sets significantly improved classification performance (F1=80.38\%). These promising results demonstrate the HRI-SA dataset’s potential to support the development of online techniques for the automatic monitoring of operators’ PSALs in remote HRT. 



We observed high false positive rates in unimodal PSAL classification (i.e., using eye-tracking only or contextual only features). Similar observations were noted in previous research focusing on events necessitating intervention~\cite{ali2025estimating, ratwani2010single}. Fusing eye movement and contextual features improved precision by 8.6-28.4\%, yet additional improvements are still necessary, with current precision at 72.89\%. Feature-importance analysis and sample RF decision trees we visualized (with F8 near the root) indicate that contextual features contributed most to PSAL detection, while eye-tracking features played an important role in reducing false positives. 
To improve detection performance, future work could explore additional features. Examples include eye movement features focused on individual notifications, contextual features such as the time a request was first sent to the interface, operator expertise, and the number of robots involved in the mission. 





In the PSAL detection analysis, we employed Leave-One-Group-Out (LOGO)  cross validation, unlike most related work using k-fold cross-validation~\cite{ali2025estimating, ratwani2010single}. LOGO holds out all data from one participant group in each fold, ensuring model evaluation on unseen participants~\cite{yates2023cross}. K-fold cross-validation can mix data from the same participants across training and test sets, potentially overestimating model performance~\cite{lumumba2024comparative}. Our findings thus provide a realistic estimate of performance. However, we did not use a separate hold-out test set due to data limitations. 

Beyond detecting PSAL occurrences, the HRI-SA dataset can support broader online SA analyses, including CSAL detection and the estimation of PSAL and CSAL values. Our participants highlighted how such online SA assessment capabilities could enable SAL-aware user adaptations, consistent with previous findings~\cite{10.1145/3743693}. They noted that user-adaptive alerts—e.g., ``there is an object to verify'' (P7, P11), ``robot 1 needs help'' (P10, P14)—can be issued when SALs are experienced, aiding operators to decide ``when to shift attention'' (P25), especially when overseeing ``multiple robots'' (P8). ``Prioritizing $<$such$>$ alerts'' (P30) could benefit from real-time SAL value estimation. Moreover, robots' awareness of operator SAL can inform adaptive actions (P6, P10, P13); e.g., when uncertain, a robot can either pause and wait for assistance (P6, P13) or proceed with notification, considering SAL experienced by the operator (P10). Participants also noted that robotic interfaces often issue excessive alerts solely based on contextual data (e.g., whether a robot is experiencing an error) even when operators are focused on relevant tasks, leading to alert fatigue, frustration, and oversight of critical alerts (P9-10). Our results also show that contextual-only models produced numerous false positives in PSAL classification. Eye-tracking features could help address these unnecessary alerts and mitigate alert fatigue.


Although the HRI-SA dataset focuses on a specific task, the assessment techniques developed with it can be translated to broad HRT contexts. For instance, the demonstrated PSAL detection aligns with recognizing general robot notifications. Nonetheless, evaluation across diverse HRT tasks is necessary to improve generalizability. HRI-SA also serves as a template for future datasets by integrating multimodal behavioral, contextual, and physiological data with ground-truth SA labels. A community effort in collecting additional datasets is needed to advance online SA analysis for HRT.



\section{Conclusion}

To enable online SA assessment, this paper introduced HRI-SA, a novel multimodal dataset in a realistic simulated HRT context. It serves as a benchmark for computational analysis of SA, benefiting HRT research. Additionally, our work is the first to evaluate Perceptual SA Latency (PSAL) throughout a dynamic HRT mission. Machine learning experiments demonstrate the feasibility of detecting (recall=88.91\%, F1=67.63\%) PSAL using generic eye-tracking features combined with limited contextual features. Integrating eye-tracking features was crucial to reduce high false positives associated with purely contextual detection. Future evaluations using the HRI-SA dataset could further explore the potential for online SA gap assessment to enable SA-aware user-adaptive robot behaviors and supervisory interfaces that support maintaining operator SA.

\section*{ACKNOWLEDGMENT}

We thank Fletcher Talbot for the guidance in developing the conditions of the simulated environment, and participants for their valuable time and insights. 



\bibliographystyle{IEEEtran}
\bibliography{references.bib}

\end{document}